\documentclass[sigconf, nonacm]{acmart}
\AtBeginDocument{%
  \providecommand\BibTeX{{%
    \normalfont B\kern-0.5em{\scshape i\kern-0.25em b}\kern-0.8em\TeX}}}

\usepackage{amsmath,amsfonts}
\usepackage{graphicx}
\usepackage{textcomp}
\usepackage{multirow}
\usepackage{color}
\usepackage[ruled, vlined,linesnumbered]{algorithm2e}
\usepackage{epstopdf}
\usepackage{comment}
\usepackage{multirow}
\usepackage{url}
\usepackage{optidef}

\theoremstyle{definition}
\newtheorem{definition}{Definition}
\newtheorem{theorem}{Theorem}

\newcommand{\our}{\textit{LMOS}\xspace}

\begin{document}

\title{Latency-Memory Optimized Splitting of Convolution \\ Neural Networks for Resource Constrained Edge Devices}

\author{Tanmay Jain}
\email{tanmayj2020@gmail.com}
\affiliation{%
  \institution{Delhi Technological University}
  \country{India}
}

\author{Avaneesh}
\email{avaneeshmaithil@gmail.com}
\affiliation{%
  \institution{Delhi Technological University}
  \country{India}
}

\author{Rohit Verma}
\email{rv355@cam.ac.uk}
\affiliation{%
 \institution{University of Cambridge}
 \country{UK}
}

\author{Rajeev Shorey}
\email{rshorey@iitd.ac.in}
\affiliation{%
 \institution{UQIDAR, IIT Delhi}
 \country{India}
}


\begin{abstract}
With the increasing reliance of users on smart devices, bringing essential computation at the edge has become a crucial requirement for any type of business. Many such computations utilize Convolution Neural Networks (CNNs) to perform AI tasks, having high resource and computation requirements, that are infeasible for edge devices. Splitting the CNN architecture to perform part of the computation on edge and remaining on the cloud is an area of research that has seen increasing interest in the field. In this paper, we assert that running CNNs between an edge device and the cloud is synonymous to solving a resource-constrained optimization problem that minimizes the latency and maximizes resource utilization at the edge. We formulate a multi-objective optimization problem and propose the \our algorithm to achieve a Pareto efficient solution. Experiments done on real-world edge devices show that, \our ensures feasible execution of different CNN models at the edge and also improves upon existing state-of-the-art approaches.
\end{abstract}

\ccsdesc[500]{Computing methodologies~Machine learning}
\ccsdesc[500]{Computing methodologies~Cooperation and coordination}

\keywords{edge computing, CNNs, multi-objective optimization}


\maketitle

\section{Introduction}~\label{introduction}
Recent times have seen an increasing trend towards bringing computation to the edge in order to increase the level of automation at the edge and obtain more realistic real-time solutions. As per a 2018 study, the percentage of data processed on the edge in one way or another would increase to $75\%$ in 2025 from a mere $10\%$ in 2018~\cite{gartner}. AI/ML on edge is linked to crucial future applications. Highly reliant on Convolution Neural Networks (CNNs)~\cite{shi,lecun1999object}, these applications include quality control in industries, facial recognition, health care, smart retail, and autonomous vehicles, to name a few. Furthermore, AI in edge computing is going to be seen in $50\%$ of all edge computing applications by 2025~\cite{idc}.

Convolution Neural Networks (CNNs)~\cite{lecun1999object} are a class of deep neural networks that utilize convolution instead of matrix multiplication in at least one of the network layers. In addition to the Convolution layers, the CNN relies on the Pooling, Rectified Linear Unit (ReLU), and Fully connected layers to provide the output volume. While computation at all of these layers is usually resource-intensive; it is to be noted that a majority of the edge devices are resource-constrained with limited memory and computational capability. This renders running CNNs on edge devices quite challenging. Existing techniques have tried to address this challenge. One class of work tries to compress the network architecture to build compact CNN models that could run on an edge device~\cite{hu2020fast,gamanayake2020cluster}. These models compromise on accuracy and many-a-time are model specific. Another class of work splits the input into smaller inputs to minimize the memory requirements~\cite{jin2019split}. These approaches are limited by the level of image splitting and would thus be model-specific. Instead of carrying out all the computations on the edge device, the other set of solutions split the CNN architecture to perform part of the processing on the cloud~\cite{mehta2020deepsplit, tang2020joint}. However, these works either split the CNN based on some model-specific empirical thresholds or rely on latency optimization, which in most cases would defer the splitting. When closely observed, the splitting of a CNN between the edge and the cloud could be formulated as a resource-constrained optimization problem that tries to minimize the latency, while also trying to maximize resource utilization on the edge device.

\begin{figure*}[!ht]
  \centering
  \includegraphics[width=1\linewidth]{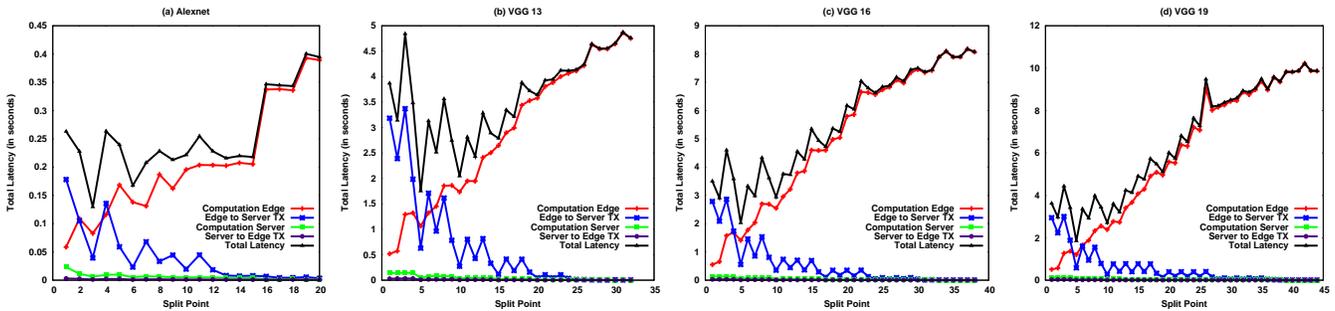}
  \caption{Total latency vs splitting point of CNN for different models.}
  \label{fig:splitmotiv}
\end{figure*}

In this paper, we formulate a multi-objective optimization problem that considers both latency and memory utilization when finding the optimal layer for splitting the CNN between the edge device and the cloud server~\footnote{We use Cloud, Server and Cloud Server interchangeably throughout the paper}. Following this, we develop the \textit{Latency-Memory Optimized Splitting} (\our) algorithm, that calculates the Pareto optimal solution for the multi-objective optimization problem. We evaluate \our over a prototype edge environment set up using Raspberry Pi4~\cite{rpi} modules. Experiments show that \our computes the optimal split point for splitting the CNN in order to minimize the latency and maximize the resource utilization without compromising on the accuracy. Further, the \our algorithm also improves upon existing solutions.

In the following section ($\S\ref{relatedwork}$), we give a brief overview of existing solutions and their limitations. Next, we define and formulate the problem ($\S\ref{problemdefinition}$) before describing \our in detail ($\S\ref{optimizationalgorithm}$). We then evaluate \our in different scenarios ($\S\ref{evaluation}$) and finally conclude with a discussion of future work ($\S\ref{conclusion}$).
\section{Related Work}~\label{relatedwork}

Convolution Neural Networks (CNN)~\cite{lecun1999object} were introduced as image recognition neural networks, but have become crucial to several computer vision and other related applications. This is evident from the numerous CNN models developed in recent times~\cite{krizhevsky2012imagenet,simonyan2014very,sandler2018mobilenetv2,szegedy2017inception}. However, when trying to run CNNs on an edge device, the high computation and memory requirements become a bottleneck~\cite{khalil2021deep}.

With the increasing need to bring AI on the edge, many works have applied different approaches to overcome this bottleneck. The naive approach is to offload all computation to the server~\cite{huang2017deep}, undermining the utility of the edge devices that could themselves perform some computation. One popular approach is to develop lightweight strategies that reduce the computation and memory requirements manifold~\cite{hu2020fast,gamanayake2020cluster,sandler2018mobilenetv2,iandola2016squeezenet,louis2019towards}. These approaches achieve computational and memory efficiency by either compressing the CNN model~\cite{iandola2016squeezenet, sandler2018mobilenetv2, hu2020fast} or using lightweight libraries~\cite{louis2019towards}. Network pruning is another approach that reduces the complexity by pruning redundant and non-informative weights~\cite{fan2019cscc,louis2019towards,hassibi1993second,han2015deep}. However, both model compression and network pruning compromise accuracy and are specific for a certain model or certain class of models~\cite{khalil2021deep}.

Another class of approaches splits the CNN architecture between the edge device and the server~\cite{mehta2020deepsplit, matsubara2019distilled, tang2020joint, zhou2019distributing, leroy2021optimal}. The algorithm-based splitting methods decide the splitting point either based on a model-specific threshold~\cite{mehta2020deepsplit} or input-output dimension size~\cite{matsubara2019distilled}. However, these are model-specific splitting and cannot be generalized. Other approaches split the CNN by trying to optimize the computation latency~\cite{tang2020joint, zhou2019distributing} limiting the utility of the edge device.

In contrast to the existing state-of-the-art approaches, there is a need for a dynamic splitting approach, that not only optimizes latency but also optimizes utilization of the edge device without compromising accuracy.

\section{Problem Definition}~\label{problemdefinition}
As discussed earlier, the problem of deciding where to split the CNN architecture could be seen as a resource-constrained optimization problem. Several system parameters have an impact on the optimization decision. In this section, we describe how and why the system parameters are considered in formulating the problem.

\subsection{Pilot Study}

When splitting a CNN between an edge device and the cloud, numerous system parameters need to be considered. We perform a set of experiments to identify these system parameters. The experiment is set up between a Raspberry Pi4 module~\cite{rpi}, used as an edge device, and an Ubuntu 20.04 system~\cite{ubuntu} as the cloud server. The RPi4 module has 16 GB storage, 4 GB RAM, and a quad-core 1.5 GHz processor. The cloud server has 8 GB RAM and an octa-core 1.5 GHz processor. Both RPi4 module and the cloud have a network connection of 10 Mbps between them. We run our experiments with four pre-trained CNN models, Alexnet (21 layers)~\cite{krizhevsky2012imagenet}, VGG13 (32 layers), VGG16 (38 layers), and VGG19 (44 layers)~\cite{simonyan2014very}. All models perform image classification based on an input image. Each model is split at different layers in different runs and the computation and transmission times are logged for each run.

In Figure~\ref{fig:splitmotiv}, we plot the total latency for each model when split at a particular layer. In addition to the total latency, we also plot the four contributory latency factors, viz., computation time at the edge device, transmission time from edge to the server, computation time at the server, and transmission time from the server to the edge device. As is evident from all the four plots in Figure~\ref{fig:splitmotiv}, the computation time at the edge device and the transmission time from the edge to the cloud are the primary contributing factors for total latency. It can be observed that splitting at different layers affects the contributing latency factors differently. Moreover, the edge device can also perform low latency computation when trying to compute more layers. Hence, it is essential to allow the edge device to perform the maximal computation in addition to minimizing the latency. Computation time at the edge, which plays a crucial part in the total latency calculation, is impacted by the edge device's computational capability. Further, the computational capability of the server also has a role in the total latency computation that impacts the splitting point. Finally, bandwidth is an important parameter to consider when computing the transmission time between edge and server. 

\subsection{Problem description}
The above set of experiments show that we not only need to minimize the latency but also maximize the computation at the edge which could be linked to the memory usage at the edge device. Therefore, we need to define two objective functions for the optimization problem, one addressing latency and the other addressing memory usage at the edge.

We assume that there are $\mathbb{L}$ layers in the CNN. After splitting the CNN there are $x_1$ layers at the edge and $x_2$ layers at the server. The memory usage, denoted by $\mathbb{X}_{edge}$, is computed as the memory required when the edge device is performing convolution over the $x_1$ layers. This forms the basis for the objective function that attempts to maximize the memory usage at the edge device.

As observed in Figure~\ref{fig:splitmotiv}, there are four components involved when considering the overall latency of the computation. However, the transmission time from server to edge is constant and negligible since the server sends a fixed low size classification output to the edge and the bandwidth is usually constant with minimal variance. This can be seen in Figure~\ref{fig:splitmotiv}. Thus, we do not include the transmission time from server to edge in the objective function. The other three latency values are considered when defining the objective function. We describe the latency components in the following sub-sections.

\subsubsection{Edge Convolution Latency ($\mathbb{T}_{edge}$):} This is the time that the edge device takes to compute the $x_1$ convolution layers. This is represented as:
\begin{equation}
    \mathbb{T}_{edge} = \frac{\mathbb{X}_{edge}|x_1}{\mathcal{C}_{edge} * \mathcal{S}_{edge}}
\end{equation}
where, $\mathbb{X}_{edge}|x_1$ is the amount of local computation done at the edge device given it has to compute $x_1$ layers. This value is computed based on the network depth, the width and the height of the kernel at each layer~\cite{giro2016memory}. The denominator defines the computational capacity of the edge device which we take as the product of the number of CPU cores $\mathcal{C}_{edge}$ and the processor speed $\mathcal{S}_{edge}$.

\subsubsection{Edge to Server Transmission Latency ($\mathbb{T}_{TX}$): } This is the transmission time for sending the intermediate results from the edge to the cloud server. $\mathbb{T}_{TX}$ is dependent on the size of the intermediate output, which, in turn, is a function of the kernel weights and the network depth at the last layer to be computed and the type of computation, i.e., Convolution, Regularization or Pooling~\cite{giro2016memory}. Given that the intermediate output size is $\mathbb{U}_{edge}|x_1$ for $x_1$ layers, and the bandwidth between the edge and server is $\mathcal{B}$, the transmission latency is computed as:
\begin{equation}
    \mathbb{T}_{TX} = \frac{\mathbb{U}_{edge}|x_1}{\mathcal{B}}
\end{equation}

\subsubsection{Server Convolution Latency ($\mathbb{T}_{server}$):} The computation latency at the server is a function of the amount of local computation performed with $x_2$ layers ($\mathbb{X}_{server}|x_2$) at the server, the number of CPU cores ($\mathcal{C}_{server}$) and the processor speed ($\mathcal{S}_{server}$). $\mathbb{T}_{server}$ is calculated as follows:
\begin{equation}
    \mathbb{T}_{server} = \frac{\mathbb{X}_{server}|x_2}{\mathcal{C}_{server} * \mathcal{S}_{server}}
\end{equation}

The above three latency components form the basis for the objective function that attempts to minimize the latency.

\subsection{Problem formulation}
We formulate the objective functions that define the optimization problem along with the related constraints as follows:

\begin{equation}
    f_1(x_1, x_2) = \frac{\mathbb{X}_{edge}|x_1}{\mathcal{C}_{edge} * \mathcal{S}_{edge}} +
    \frac{\mathbb{U}_{edge}|x_1}{\mathcal{B}} + 
    \frac{\mathbb{X}_{server}|x_2}{\mathcal{C}_{server} * \mathcal{S}_{server}}
    \label{eqn:f1}
\end{equation}

\begin{equation}
    f_2(x_1) = \mathbb{X}_{edge}|x_1
    \label{eqn:f2}
\end{equation}

Equation~\ref{eqn:f1} defines the latency objective function that is the end-to-end latency given that the edge computes $x_1$ layers and the server computes $x_2$ layers of the CNN. Equation~\ref{eqn:f2} defines the memory objective function that is computed given the $x_1$ edge layers. The optimization problem can thus be represented as follows:

\begin{mini}|s|
{}{\mathbb{F} = (f_1, -f_2)}
{}{}
\addConstraint{\mathbb{X}_{edge}|x_1 \leq \mathbb{M}}
\addConstraint{x_1 + x_2 = \mathbb{L}}{}
\addConstraint{0 \leq x_1 \leq \mathbb{L}}{}
\addConstraint{0 \leq x_2 \leq \mathbb{L}}{}
\label{eqn:f3}
\end{mini}

With equations~\ref{eqn:f1} and~\ref{eqn:f2}, we wish to minimize $f_1$ and maximize $f_2$. We formulate the problem as in Equation~\ref{eqn:f3} where we minimize $f_1$ and $-f_2$. The multi-objective optimization problem must adhere to the four constraints. First, the local computation memory required on the edge device must not exceed the total available storage ($\mathbb{M}$) at the edge device. Second, the sum of the number of layers at the edge and the server should always add up to the total layers of the CNN ($\mathbb{L}$). Finally, the number of layers at the edge and the server should not be negative or exceed $\mathbb{L}$.
\section{The \our Optimization Algorithm}\label{optimizationalgorithm}
In this section, we design the \textit{Latency-Memory Optimized Splitting} (\our) algorithm to obtain an optimal solution for the multi-objective optimization problem $\mathbb{F}$. We first define a few terms.

\begin{definition}[\textit{Solution Space}]
$\mathcal{X}$ is the set of feasible solutions that $\mathbb{F}$ can have. Hence, any solution vector $\Vec{x} = \{x_1, x_2\} \in \mathcal{X}$. 
\end{definition}

\begin{definition}[\textit{Objective Space}]
We represent any evaluation vector $\mathbb{F}(\Vec{x})$ as $\Vec{y}$ for a given solution vector $\Vec{x}$. Then the objective space is defined as $\mathcal{Y} = \{ \Vec{y} : y_i = f_i(\Vec{x}), \forall \Vec{x} \in \mathcal{X}, i \in [1,2] \}$
\end{definition}

An optimal solution vector $\Vec{x}_{opt}$ is said to be the one which dominates all other $\Vec{x} \in \mathcal{X} - \Vec{x}_{opt}$. We define dominance as follows;
\begin{definition}[\textit{Dominance}]
Let $\Vec{y}$ and $\Vec{y}'$ be two evaluation vectors in $\mathcal{Y}$. We say $\Vec{y}$ dominates $\Vec{y}'$, if and only if $y_1 \leq y'_1$ and $y_2 \leq y'_2$. Further, at least one inequality should be strict.
\end{definition}

We observe from Figure~\ref{fig:splitmotiv} that the total latency is generally lower when splitting is done at lower layers while the computation latency at the edge is higher when splitting is done at higher layers. Higher computation latency at the edge implies more computation is performed at the edge leading to higher edge memory usage, thus implying that memory usage is higher when splitting is done at the higher layers. Hence, a dominant solution for the optimization problem cannot be obtained and there is a need to compute a non-dominant or \textit{Pareto-efficient}~\cite{debreu1954valuation} solution for $\mathbb{F}$.

We develop an algorithm based on the \textit{$\epsilon$-constrained method}~\cite{chankong2008multiobjective} to solve the multi-objective optimization problem. We find the optimal solution for one of the objective functions and represent the other as a constraint. Thus, we rewrite Equation~\ref{eqn:f3} as follows:

\begin{mini}|s|
{}{\mathbb{F}_i(\epsilon_j) = (f_i)}
{}{}
\addConstraint{f_j \leq \epsilon_j, i,j \in [1,2] \land i \neq j}
\addConstraint{\mathbb{X}_{edge}|x_1 \leq \mathbb{M}}
\addConstraint{x_1 + x_2 = \mathbb{L}}{}
\addConstraint{0 \leq x_1 \leq \mathbb{L}}{}
\addConstraint{0 \leq x_2 \leq \mathbb{L}}{}
\label{eqn:f4}
\end{mini}

The solution of Equation~\ref{eqn:f4} is based on the following theorems\footnote{Proof of the theorems is out of scope of this paper and is available in the literature~\cite{chankong2008multiobjective}}~\cite{chankong2008multiobjective};
\begin{theorem}
$\Vec{x}_e$ is an efficient solution of $\mathbb{F}$ if and only if $\Vec{x}_e$ solves $\mathbb{F}_i(\epsilon_j)$ for $i \in [1,2]$
\end{theorem}

\begin{theorem}
If $\Vec{x}_e$ solves $\mathbb{F}_i(\epsilon_j)$ for some $i$ and if this is a unique solution, then $\Vec{x}_e$ is an efficient solution for $\mathbb{F}$.
\end{theorem}

The two theorems summarize that an exact Pareto front can be found for $\mathbb{F}$ by solving the \textit{$\epsilon$-constrained} problems, given that we get a solution for every point in $\mathcal{F}$, which is the Pareto front.

\begin{definition}[\textit{Pareto Front}]
$\mathcal{F} = \{ \mathbb{F}(\Vec{x}): \Vec{x}$ is Pareto efficient in $\mathcal{X} \}$
\end{definition}

\begin{algorithm}
\SetAlgoLined
\KwResult{The Pareto Optimal Solution $\Vec{x}_e$ for $\mathbb{F}$}
 $i \gets 1, j \gets 2$ or $i \gets 2, j \gets 1$\;
 
 $\Vec{y}_{ideal} \gets \{ \min\limits_{\Vec{y} \in \mathcal{Y}} y_1, \min\limits_{\Vec{y} \in \mathcal{Y}} y_2 \}$\; 
 
 $\Vec{y}_{nadir} \gets \{ \min\limits_{\Vec{y} \in \mathcal{Y}} \{ y_1: y_2 \in \Vec{y}_{ideal}\}, \min\limits_{\Vec{y} \in \mathcal{Y}} \{ y_2: y_1 \in \Vec{y}_{ideal}\} \}$\;
 $\epsilon_j \gets y_j \in \Vec{y}_{nadir}$\;
 \While{$\epsilon_j \geq y_j \in \Vec{y}_{ideal}$}{
  Solve $\mathbb{F}_i(\epsilon_j)$ to get $(y_i, y_j)$\;
  $\mathcal{F} += (y_i, y_j)$\;
  $\epsilon_j \gets$ next best value of $y_j$ \;
 }
 Remove dominant points in $\mathcal{F}$ if any\;
 Return $\Vec{x}_e$ from $\mathcal{F}$ which minimizes $f_i(\Vec{x})$
 \caption{The \our Optimization Algorithm}
 \label{algo:optimization}
\end{algorithm}

In Algorithm~\ref{algo:optimization}, we describe how the Pareto optimal solution is obtained for $\mathbb{F}$ using the $\epsilon$-constrained method. The first task is to decide which objective function is to be optimized. Consequently, the other objective function will be constrained. This sets the value of $i$ and $j$. Subsequently, the \textit{ideal} and \textit{nadir} points are obtained for both the objective functions. As the aforementioned theorems state, solving multiple $\epsilon$-constrained problems would give the efficient solution by creating the Pareto front. The algorithm achieves this by setting $\epsilon_j$ as the worst possible value first and then decreasing it to reach the ideal value. For every value of $\epsilon_j$, the algorithm solves $\mathbb{F}_i(\epsilon_j)$, and the solution is added to $\mathcal{F}$. $\epsilon_j$ is then set at the next best value in $\mathcal{Y}$. Once the loop is over, the algorithm checks if there are any dominant points in $\mathcal{F}$ and removes these since we are only interested in the non-dominating points. It should be noted that the dominating points are only dominating in $\mathcal{F}$ and hence do not follow the dominance rule defined earlier. Finally, based on the ranking method, the Pareto optimal solution is chosen as the one which minimizes $f_i(\Vec{x})$.

In this paper, we use the latency objective function ($f_1$) as a constraint and optimize the memory requirement function ($f_2$). Further, we solve for the optimization problem $\mathbb{F}_2(\epsilon_1)$. This decision has been taken based upon an empirical analysis, which reveals that when running \our for $\mathbb{F}_1(\epsilon_2)$, not all values of $\epsilon_2$ give a solution for $\mathbb{F}_1(\epsilon_2)$ (Please refer to Line 6 of Algorithm~\ref{algo:optimization}).
\section{Performance Evaluation}~\label{evaluation}
In this section, we first describe the experimental setup of a prototype edge environment. Subsequently, we perform sensitivity analysis of \our followed by the performance evaluation of \our.

\subsection{Experiment Setup}
We build a prototype edge-cloud set up using Raspberry Pi4 and Ubuntu server to evaluate our work. We use four Raspberry Pi4 modules as edge devices and an Ubuntu 20 system as the cloud server. The RPi4 modules have storage sizes of 16 GB, 8 GB, 4 GB, and 2 GB. However, all have the same 4 GB RAM and a quad-core 1.5 GHz processor. The cloud server is the same used in Section~\ref{problemdefinition}. It runs a Ubuntu 20.04 OS, with an 8 GB RAM and octa-core 1.5 GHz processor. The RPi4 modules and the cloud server are connected to a Wi-Fi network providing a bandwidth of 10 Mbps.
We utilize pre-trained CNN models of Alexnet (21 layers)~\cite{krizhevsky2012imagenet}, VGG13 (32 layers), VGG16 (38 layers), VGG19 (44 layers)~\cite{simonyan2014very}, and MobileNet v2 (21 layers)~\cite{sandler2018mobilenetv2} available from PyTorch Hub~\cite{torchhub} for evaluation.



\subsection{Parameter Sensitivity of \our}

\begin{table}[!ht]
\scriptsize
\begin{tabular}{|l|l|l|}
\hline
\textbf{Parameter} & \textbf{Range} & \textbf{Default Value} \\ \hline
Bandwidth          & 1 - 200 Mbps   & 10 Mbps                    \\ \hline
Edge Cores         & 1 - 8          & 2                      \\ \hline
Edge CPU Speed     & 1 - 2 GHz      & 1.5 GHz                \\ \hline
Edge Storage Size   & 256 - 16000 MB & 8000 MB                \\ \hline
Server Cores       & 1 - 8          & 8                      \\ \hline
Server CPU Speed   & 1.5 - 3.2 GHz  & 2.6 GHz                \\ \hline
\end{tabular}
\caption{Range and default value of all the parameters used in the sensitivity analysis of \our}
\label{tab:sensitivity}
\end{table}

We run \our for five scenarios, where we fix the number of layers in the CNN model to be $10, 70, 120, 170, \&$ $220$ and observe how the splitting point changes when specific parameters are varied. Table~\ref{tab:sensitivity} lists the parameters with their range and default values\footnote{Default value is the value a parameter takes when another parameter is varied.}.

\begin{figure}[!ht]
  \centering
  \includegraphics[width=1\linewidth]{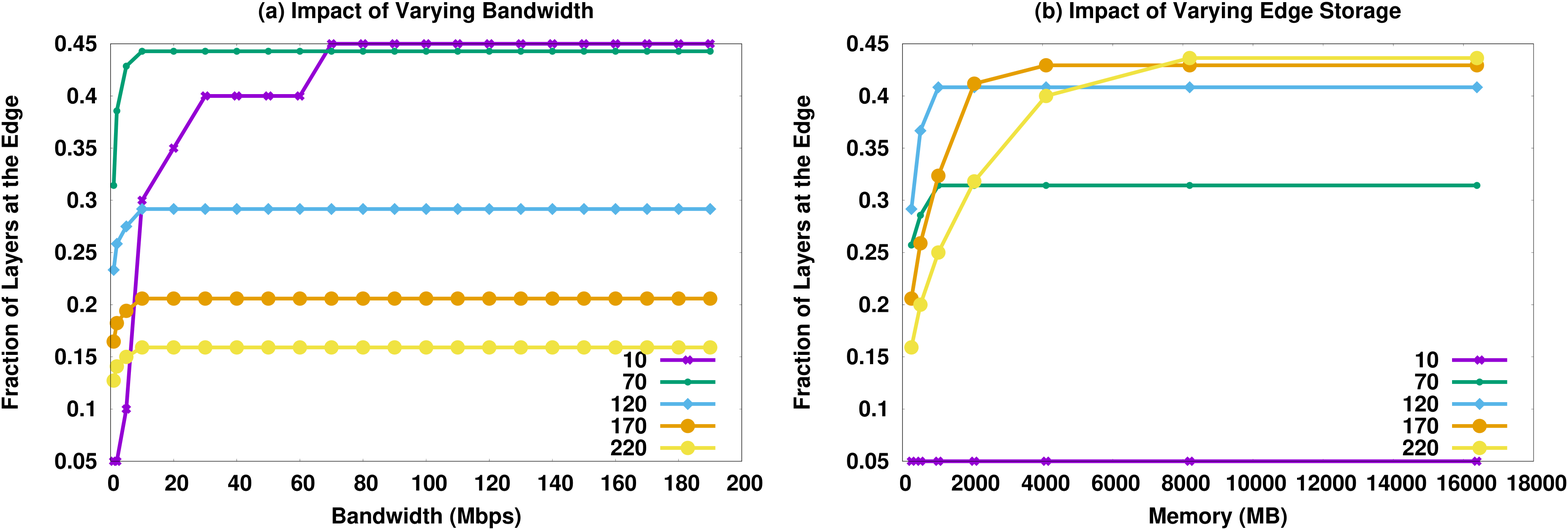}
  \caption{Impact of varying Bandwidth and Edge Storage}
  \label{fig:bw_mem}
\end{figure}


\begin{figure}[!ht]
  \centering
  \includegraphics[width=1\linewidth]{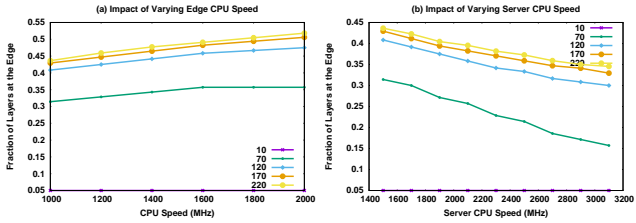}
  \caption{Impact of varying Edge and Server CPU speed}
  \label{fig:cpu}
\end{figure}

We calculate the fraction of layers computed at the edge device when we vary a particular parameter in Table~\ref{tab:sensitivity}. Figure~\ref{fig:bw_mem}(a) shows the impact of varying bandwidth and Figure~\ref{fig:bw_mem}(b) shows the impact of varying edge storage size. It is observed that at low bandwidth, \our prefers to compute less on the edge device as more computation on the edge increases the intermediate output size thus increasing the transmission time. However, after a certain value, the bandwidth does not have any impact on the CNN splitting. Similarly, at low storage size, there are fewer layers at the edge that increases considerably as the storage size increases and then becomes constant. Another interesting observation is that for $10$ layers, the storage size does not have any impact. This is primarily because of the low memory requirement in the smaller CNN and therefore the primary driving factor is latency computation.

Figure~\ref{fig:cpu} shows the impact of computation capability of the edge device and server on the optimization results. Intuitively, increasing the computational capability by increasing the CPU speed at the edge increases the number of layers that the edge device can compute, while, the reverse happens when the computational capability of the server is increased. We obtain similar results when varying the number of cores at the edge and the server.

\subsection{Performance Evaluation of \our}
We now evaluate how splitting the CNN impacts the computation of the CNN models between the edge device and the cloud server. We show the impact of splitting on the accuracy, latency and memory utilization when varying the bandwidth between the RPi4 modules and cloud server and varying the storage size of the RPi4 modules. We run the experiments for four different models, viz., AlexNet, VGG13, VGG16, and VGG19. We evaluate all models with $100$ input images and the reported results are averaged over the $100$ runs.

\begin{figure}[!ht]
  \centering
  \includegraphics[width=1\linewidth]{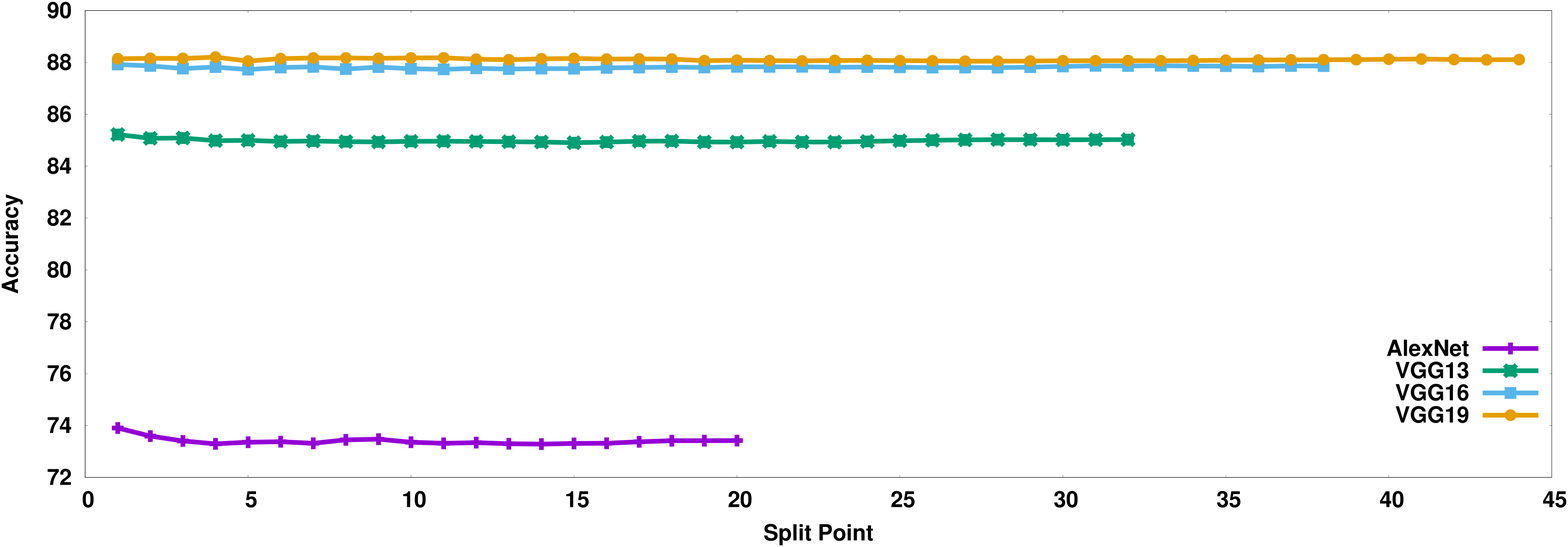}
  \caption{Impact of splitting the CNN on Model Accuracy}
  \label{fig:accuracy}
\end{figure}

In our results, we refer to the \textit{classification accuracy}\cite{cnnaccuracy} as the accuracy of the model. The \textit{classification accuracy} is defined as the ratio of correct predictions to the total number of input samples. In order to show that splitting has no impact on the accuracy of the model, we split the models at all layers and compute the result. As can be seen in Figure~\ref{fig:accuracy}, splitting does not affect the accuracy of the models. This strengthens the fact that splitting the CNN models between the edge and the server does not impact the model output.


\begin{table}[!ht]
\scriptsize
\begin{tabular}{|l|l|l|l|l|l|l|l|l|}
\hline
\multirow{2}{*}{\textbf{Model}} & \multicolumn{4}{l|}{\textbf{Bandwidth (Mbps)}}       & \multicolumn{4}{l|}{\textbf{Storage (GB)}}         \\ \cline{2-9} 
                                 & \textbf{0.5} & \textbf{1} & \textbf{5} & \textbf{10} & \textbf{2} & \textbf{4} & \textbf{8} & \textbf{16} \\ \hline
\textbf{AlexNet}                 & 8            & 18         & 19         & 19          & 10         & 13         & 17         & 19          \\ \hline
\textbf{VGG13}                   & 10           & 18         & 23         & 30          & 10         & 13         & 17         & 21          \\ \hline
\textbf{VGG16}                   & 10           & 18         & 23         & 32          & 10         & 13         & 17         & 21          \\ \hline
\textbf{VGG19}                   & 10           & 18         & 23         & 32          & 11         & 13         & 17         & 21          \\ \hline
\end{tabular}
\caption{Number of CNN layers computed at edge when varying bandwidth and varying storage size of edge device}
\label{tab:bandwidthpi}
\end{table}

\begin{figure}[!ht]
  \centering
  \includegraphics[width=1\linewidth]{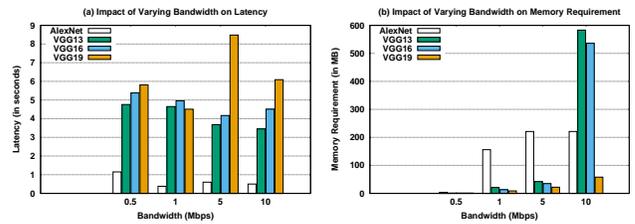}
  \caption{Impact of bandwidth variation}
  \label{fig:bw_mem_latency}
\end{figure}

We vary the bandwidth between the 16 GB RPi4 module and the cloud server to show the change in split points and the resulting memory and latency values for these scenarios. The number of layers computed at the edge device is shown in Table~\ref{tab:bandwidthpi}. In Figure~\ref{fig:bw_mem_latency}, we show the impact of bandwidth variation on latency and memory usage. The key takeaway for the latency result is that an increase in bandwidth does not always reduce the latency. For example, the latency increases for VGG19 from 1 Mbps to 5 Mbps. Such an increase could be linked to the fact that an increase in bandwidth implies that the edge can send more data to the server and hence can compute more layers. This leads to more computation at the edge device resulting in higher convolution time at the edge, thus increasing the total latency. However, an increase in bandwidth always leads to an increase in memory requirement which is due to the increase in the number of layers being computed at the edge device. VGG13 and VGG16 show a higher increase since almost all the layers of the CNN are computed at the edge device thus requiring more memory.


\begin{figure}[!ht]
  \centering
  \includegraphics[width=1\linewidth]{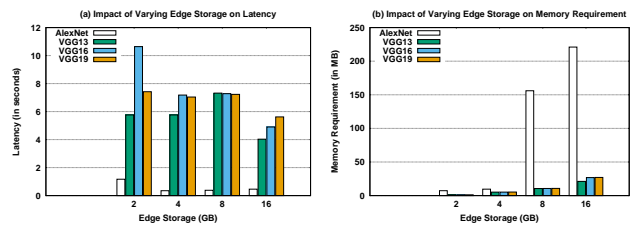}
  \caption{Impact of RPi storage variation}
  \label{fig:storage_mem_latency}
\end{figure}

We show a similar analysis by varying the storage size of the RPi4 modules. The corresponding layers at the edge are given in Table~\ref{tab:bandwidthpi}. Figure~\ref{fig:storage_mem_latency} shows the impact on latency and memory requirement by varying the storage size at the edge. Increasing the storage leads to more computation at the edge and thus leads to higher latency; however, this need not always be true. The decrease in latency could be linked to the type of computations the edge performs (Convolution, Regularization or Pooling) since Convolution takes more time than Regularization or Pooling. It is to be noted that the memory requirement always increases with increasing storage.

\subsection{Competing Approaches}
We compare \our with four competing approaches, one is based on latency optimization, two are boundary approaches and the final is a random approach.

\subsubsection{Latency Optimized Approach (LOA): } Several works~\cite{kang2017neurosurgeon,li2018edge} split the layers to optimize the latency of the system. Hence, there isn't any optimization on utilization of the edge device.

\subsubsection{Edge Computation Only (ECO):} In this approach, CNN computation is done at the edge device only with no server interaction.

\subsubsection{Server Computation Only (SCO):} The entire CNN computation is done at the server, the edge device is only responsible to send the input to the server.

\subsubsection{Random Splitting (RS):} A random number is generated for each trial and the CNN is split at that layer.

\begin{table}[!ht]
\scriptsize
\begin{tabular}{|l|l|l|l|l|}
\hline
\textbf{Approach} & \textbf{AlexNet} & \textbf{VGG13} & \textbf{VGG16} & \textbf{VGG19} \\ \hline
\textbf{\our}       & 19                & 21             & 21             & 21             \\ \hline
\textbf{LOA}         & 1               & 1             & 3             & 7             \\ \hline
\textbf{ECO}         & 21               & 32             & 38             & 44             \\ \hline
\end{tabular}
\caption{Number of layers at edge for competing approaches. SCO has zero layers on edge and RS generates random layers.}
\label{tab:compete}
\end{table}

\vspace{-2em}

\begin{figure}[!ht]
  \centering
  \includegraphics[width=1\linewidth]{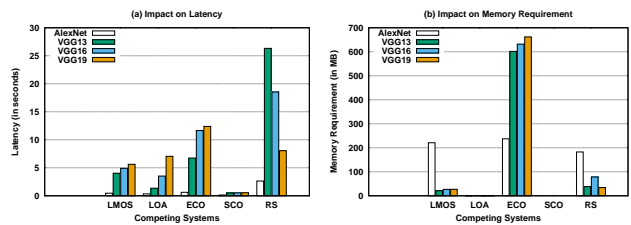}
  \caption{Comparing \our with other approaches}
  \label{fig:competing}
\end{figure}

We run the above approaches for the four CNN models between the 16 GB RPi4 module and the server. We evaluate all CNN models with $100$ input images and the reported results are averaged over the $100$ runs. The bandwidth between the edge device and server is 10 Mbps. Table~\ref{tab:compete} shows the layers computed at the edge for all the models except SCO and RS. There are no layers at the edge for SCO and different split layers are generated for RS in each trial. 

The average values are reported in Figure~\ref{fig:competing}. We observe that as expected, RS shows promising results in some scenarios but poor results in others and therefore isn't a stable approach. SCO has the minimum latency but also utilizes negligible edge memory which is undesirable. On the other hand, ECO consumes maximum memory but has higher latency than the other approaches. Although LOA has low latency than LMOS for all the models, memory utilization is very low, almost the same as SCO. In contrast to these systems, \our has latency requirements comparable to LOA but has higher memory utilization, making it a better alternative than the other competing algorithms.

\subsection{Comparison with Edge Optimized Model}
We compare \our with MobileNetv2~\cite{sandler2018mobilenetv2} which is optimized for edge devices. Instead of comparing MobileNetV2 with all four CNN models, in this experiment we use VGG19, that has the highest number of layers and the maximum memory requirements. Furthermore, VGG19 also has the highest accuracy than the other models. Hence, establishing that VGG19 with \our gives better results than MobileNetV2 would help us prove that splitting is a better alternative than compressing a CNN model. We run MobileNetv2 on the RPi4 module and run the VGG19 model both with and without \our. We evaluate both models with $100$ input images and the reported results are averaged over the $100$ runs.

\begin{table}[!ht]
\scriptsize
\begin{tabular}{|l|l|l|l|}
\hline
\textbf{Model}     & \textbf{Accuracy} & \textbf{Latency} & \textbf{Memory Used} \\ \hline
\textbf{VGG19 No Split} & 0.91              & 12.39 s           & 662 MB                    \\ \hline
\textbf{MobileNetV2}  & 0.83              & 1.3 s             & 29.77 MB                  \\ \hline
\textbf{VGG19 with LMOS}         & 0.91              & 5.62 s            & 27.03 MB                  \\ \hline
\end{tabular}
\caption{Comparing with edge optimized model}
\label{tab:mobilenet}
\end{table}

\vspace{-2em}

Comparison results for the three models are given in Table~\ref{tab:mobilenet}. While VGG19 gives better accuracy, due to CNN splitting, memory utilization is less than that of MobileNetV2 for VGG19 with \our. Furthermore, the total latency when using LMOS is $\approx 4$  seconds greater than that of MobileNetV2 which is a small trade-off for improving accuracy. The above result strengthens our assertion that splitting is a superior alternative for running CNN at the edge.
\section{Conclusion}~\label{conclusion}
In this paper, we show that running resource-intensive CNNs at the edge could be formulated as a multi-objective optimization problem that minimizes the end-to-end latency and maximizes the memory utilization at the edge device. We have proposed \our -- an $epsilon$-constrained algorithm to solve the optimization problem. Our experiments performed on a prototype edge environment show that \our provides an optimal solution for splitting the CNN. The key takeaways from the paper are: (i) splitting the CNN does not impact the model accuracy, (ii) \our is a better alternative as compared to the existing splitting-based approaches since it ensures that both total latency and memory utilization is optimized, (iii) splitting a CNN with \our is a superior alternative for running CNN at the edge as compared to other edge-optimized CNN models that only run on the edge device.

There exist key aspects that need further analysis in order to improve \our. For instance, energy consumption is a crucial parameter when considering edge devices. Running resource-hungry applications is likely to drain the device power. Thus, including the energy metric in the optimization problem could be an important extension. Further, it is important to investigate whether \our could be generalised to other neural networks. Yet another direction of investigation could be the use of smartphones as edge devices with additional constraints.

\bibliographystyle{ACM-Reference-Format}
\bibliography{ref}


\begin{thebibliography}{32}


\ifx \showCODEN    \undefined \def \showCODEN     #1{\unskip}     \fi
\ifx \showDOI      \undefined \def \showDOI       #1{#1}\fi
\ifx \showISBNx    \undefined \def \showISBNx     #1{\unskip}     \fi
\ifx \showISBNxiii \undefined \def \showISBNxiii  #1{\unskip}     \fi
\ifx \showISSN     \undefined \def \showISSN      #1{\unskip}     \fi
\ifx \showLCCN     \undefined \def \showLCCN      #1{\unskip}     \fi
\ifx \shownote     \undefined \def \shownote      #1{#1}          \fi
\ifx \showarticletitle \undefined \def \showarticletitle #1{#1}   \fi
\ifx \showURL      \undefined \def \showURL       {\relax}        \fi
\providecommand\bibfield[2]{#2}
\providecommand\bibinfo[2]{#2}
\providecommand\natexlab[1]{#1}
\providecommand\showeprint[2][]{arXiv:#2}

\bibitem[\protect\citeauthoryear{Chankong and Haimes}{Chankong and
  Haimes}{2008}]%
        {chankong2008multiobjective}
\bibfield{author}{\bibinfo{person}{Vira Chankong} {and}
  \bibinfo{person}{Yacov~Y Haimes}.} \bibinfo{year}{2008}\natexlab{}.
\newblock \bibinfo{booktitle}{\emph{Multiobjective decision making: theory and
  methodology}}.
\newblock \bibinfo{publisher}{Courier Dover Publications}.
\newblock


\bibitem[\protect\citeauthoryear{Debreu}{Debreu}{1954}]%
        {debreu1954valuation}
\bibfield{author}{\bibinfo{person}{Gerard Debreu}.}
  \bibinfo{year}{1954}\natexlab{}.
\newblock \showarticletitle{Valuation equilibrium and Pareto optimum}.
\newblock \bibinfo{journal}{\emph{Proceedings of the National Academy of
  Sciences of the United States of America}} \bibinfo{volume}{40},
  \bibinfo{number}{7} (\bibinfo{year}{1954}), \bibinfo{pages}{588}.
\newblock


\bibitem[\protect\citeauthoryear{Fan, Yu, Lu, Jiao, Xu, Liu, and Liu}{Fan
  et~al\mbox{.}}{2019}]%
        {fan2019cscc}
\bibfield{author}{\bibinfo{person}{Shengyu Fan}, \bibinfo{person}{Hui Yu},
  \bibinfo{person}{Dianjie Lu}, \bibinfo{person}{Shuai Jiao},
  \bibinfo{person}{Weizhi Xu}, \bibinfo{person}{Fangai Liu}, {and}
  \bibinfo{person}{Zhiyong Liu}.} \bibinfo{year}{2019}\natexlab{}.
\newblock \showarticletitle{CSCC: convolution split compression calculation
  algorithm for deep neural network}.
\newblock \bibinfo{journal}{\emph{IEEE Access}}  \bibinfo{volume}{7}
  (\bibinfo{year}{2019}), \bibinfo{pages}{71607--71615}.
\newblock


\bibitem[\protect\citeauthoryear{Gamanayake, Jayasinghe, Ng, and
  Yuen}{Gamanayake et~al\mbox{.}}{2020}]%
        {gamanayake2020cluster}
\bibfield{author}{\bibinfo{person}{Chinthaka Gamanayake},
  \bibinfo{person}{Lahiru Jayasinghe}, \bibinfo{person}{Benny Kai~Kiat Ng},
  {and} \bibinfo{person}{Chau Yuen}.} \bibinfo{year}{2020}\natexlab{}.
\newblock \showarticletitle{Cluster pruning: An efficient filter pruning method
  for edge ai vision applications}.
\newblock \bibinfo{journal}{\emph{IEEE Journal of Selected Topics in Signal
  Processing}} \bibinfo{volume}{14}, \bibinfo{number}{4}
  (\bibinfo{year}{2020}), \bibinfo{pages}{802--816}.
\newblock


\bibitem[\protect\citeauthoryear{Gir{\'o}-i Nieto, Sayrol, Salvador, Torres,
  Mohedano, and McGuinness}{Gir{\'o}-i Nieto et~al\mbox{.}}{2016}]%
        {giro2016memory}
\bibfield{author}{\bibinfo{person}{X Gir{\'o}-i Nieto}, \bibinfo{person}{E
  Sayrol}, \bibinfo{person}{A Salvador}, \bibinfo{person}{J Torres},
  \bibinfo{person}{E Mohedano}, {and} \bibinfo{person}{K McGuinness}.}
  \bibinfo{year}{2016}\natexlab{}.
\newblock \bibinfo{title}{Memory usage and computational considerations}.
\newblock
\newblock


\bibitem[\protect\citeauthoryear{Google}{Google}{[n.d.]}]%
        {cnnaccuracy}
\bibfield{author}{\bibinfo{person}{Google}.} \bibinfo{year}{[n.d.]}\natexlab{}.
\newblock \bibinfo{booktitle}{\emph{Classification Accuracy}}.
\newblock
\urldef\tempurl%
\url{https://developers.google.com/machine-learning/crash-course/classification/accuracy}
\showURL{%
Retrieved July 2021 from \tempurl}


\bibitem[\protect\citeauthoryear{Han, Mao, and Dally}{Han
  et~al\mbox{.}}{2015}]%
        {han2015deep}
\bibfield{author}{\bibinfo{person}{Song Han}, \bibinfo{person}{Huizi Mao},
  {and} \bibinfo{person}{William~J Dally}.} \bibinfo{year}{2015}\natexlab{}.
\newblock \showarticletitle{Deep compression: Compressing deep neural networks
  with pruning, trained quantization and huffman coding}.
\newblock \bibinfo{journal}{\emph{arXiv preprint arXiv:1510.00149}}
  (\bibinfo{year}{2015}).
\newblock


\bibitem[\protect\citeauthoryear{Hassibi and Stork}{Hassibi and Stork}{1993}]%
        {hassibi1993second}
\bibfield{author}{\bibinfo{person}{Babak Hassibi} {and}
  \bibinfo{person}{David~G Stork}.} \bibinfo{year}{1993}\natexlab{}.
\newblock \bibinfo{booktitle}{\emph{Second order derivatives for network
  pruning: Optimal brain surgeon}}.
\newblock \bibinfo{publisher}{Morgan Kaufmann}.
\newblock


\bibitem[\protect\citeauthoryear{Hu and Krishnamachari}{Hu and
  Krishnamachari}{2020}]%
        {hu2020fast}
\bibfield{author}{\bibinfo{person}{Diyi Hu} {and} \bibinfo{person}{Bhaskar
  Krishnamachari}.} \bibinfo{year}{2020}\natexlab{}.
\newblock \showarticletitle{Fast and accurate streaming CNN inference via
  communication compression on the edge}. In \bibinfo{booktitle}{\emph{2020
  IEEE/ACM Fifth International Conference on Internet-of-Things Design and
  Implementation (IoTDI)}}. IEEE, \bibinfo{pages}{157--163}.
\newblock


\bibitem[\protect\citeauthoryear{Huang, Ma, Fan, Liu, and Gong}{Huang
  et~al\mbox{.}}{2017}]%
        {huang2017deep}
\bibfield{author}{\bibinfo{person}{Yutao Huang}, \bibinfo{person}{Xiaoqiang
  Ma}, \bibinfo{person}{Xiaoyi Fan}, \bibinfo{person}{Jiangchuan Liu}, {and}
  \bibinfo{person}{Wei Gong}.} \bibinfo{year}{2017}\natexlab{}.
\newblock \showarticletitle{When deep learning meets edge computing}. In
  \bibinfo{booktitle}{\emph{2017 IEEE 25th international conference on network
  protocols (ICNP)}}. IEEE, \bibinfo{pages}{1--2}.
\newblock


\bibitem[\protect\citeauthoryear{Iandola, Han, Moskewicz, Ashraf, Dally, and
  Keutzer}{Iandola et~al\mbox{.}}{2016}]%
        {iandola2016squeezenet}
\bibfield{author}{\bibinfo{person}{Forrest~N Iandola}, \bibinfo{person}{Song
  Han}, \bibinfo{person}{Matthew~W Moskewicz}, \bibinfo{person}{Khalid Ashraf},
  \bibinfo{person}{William~J Dally}, {and} \bibinfo{person}{Kurt Keutzer}.}
  \bibinfo{year}{2016}\natexlab{}.
\newblock \showarticletitle{SqueezeNet: AlexNet-level accuracy with 50x fewer
  parameters and< 0.5 MB model size}.
\newblock \bibinfo{journal}{\emph{arXiv preprint arXiv:1602.07360}}
  (\bibinfo{year}{2016}).
\newblock


\bibitem[\protect\citeauthoryear{IDC}{IDC}{[n.d.]}]%
        {idc}
\bibfield{author}{\bibinfo{person}{IDC}.} \bibinfo{year}{[n.d.]}\natexlab{}.
\newblock \bibinfo{booktitle}{\emph{IDC FutureScape: Worldwide Analytics and
  Artificial Intelligence 2019 Predictions}}.
\newblock
\urldef\tempurl%
\url{https://www.idc.com/research/viewtoc.jsp?containerId=US44389418}
\showURL{%
Retrieved July 2021 from \tempurl}


\bibitem[\protect\citeauthoryear{Illing}{Illing}{[n.d.]}]%
        {shi}
\bibfield{author}{\bibinfo{person}{Robert Illing}.}
  \bibinfo{year}{[n.d.]}\natexlab{}.
\newblock \bibinfo{booktitle}{\emph{AI and ML in edge computing: Benefits,
  applications, and how they’re driving the future of business}}.
\newblock
\urldef\tempurl%
\url{https://blog.shi.com/next-generation-infrastructure/ai-and-ml-in-edge-computing/}
\showURL{%
Retrieved July 2021 from \tempurl}


\bibitem[\protect\citeauthoryear{Jin and Hong}{Jin and Hong}{2019}]%
        {jin2019split}
\bibfield{author}{\bibinfo{person}{Tian Jin} {and} \bibinfo{person}{Seokin
  Hong}.} \bibinfo{year}{2019}\natexlab{}.
\newblock \showarticletitle{Split-cnn: Splitting window-based operations in
  convolutional neural networks for memory system optimization}. In
  \bibinfo{booktitle}{\emph{Proceedings of the Twenty-Fourth International
  Conference on Architectural Support for Programming Languages and Operating
  Systems}}. \bibinfo{pages}{835--847}.
\newblock


\bibitem[\protect\citeauthoryear{Kang, Hauswald, Gao, Rovinski, Mudge, Mars,
  and Tang}{Kang et~al\mbox{.}}{2017}]%
        {kang2017neurosurgeon}
\bibfield{author}{\bibinfo{person}{Yiping Kang}, \bibinfo{person}{Johann
  Hauswald}, \bibinfo{person}{Cao Gao}, \bibinfo{person}{Austin Rovinski},
  \bibinfo{person}{Trevor Mudge}, \bibinfo{person}{Jason Mars}, {and}
  \bibinfo{person}{Lingjia Tang}.} \bibinfo{year}{2017}\natexlab{}.
\newblock \showarticletitle{Neurosurgeon: Collaborative intelligence between
  the cloud and mobile edge}.
\newblock \bibinfo{journal}{\emph{ACM SIGARCH Computer Architecture News}}
  \bibinfo{volume}{45}, \bibinfo{number}{1} (\bibinfo{year}{2017}),
  \bibinfo{pages}{615--629}.
\newblock


\bibitem[\protect\citeauthoryear{Khalil, Saeed, Masood, Fard, Alouini, and
  Al-Naffouri}{Khalil et~al\mbox{.}}{2021}]%
        {khalil2021deep}
\bibfield{author}{\bibinfo{person}{Ruhul~Amin Khalil}, \bibinfo{person}{Nasir
  Saeed}, \bibinfo{person}{Mudassir Masood}, \bibinfo{person}{Yasaman~Moradi
  Fard}, \bibinfo{person}{Mohamed-Slim Alouini}, {and} \bibinfo{person}{Tareq~Y
  Al-Naffouri}.} \bibinfo{year}{2021}\natexlab{}.
\newblock \showarticletitle{Deep Learning in the Industrial Internet of Things:
  Potentials, Challenges, and Emerging Applications}.
\newblock \bibinfo{journal}{\emph{IEEE Internet of Things Journal}}
  (\bibinfo{year}{2021}).
\newblock


\bibitem[\protect\citeauthoryear{Krizhevsky, Sutskever, and Hinton}{Krizhevsky
  et~al\mbox{.}}{2012}]%
        {krizhevsky2012imagenet}
\bibfield{author}{\bibinfo{person}{Alex Krizhevsky}, \bibinfo{person}{Ilya
  Sutskever}, {and} \bibinfo{person}{Geoffrey~E Hinton}.}
  \bibinfo{year}{2012}\natexlab{}.
\newblock \showarticletitle{Imagenet classification with deep convolutional
  neural networks}.
\newblock \bibinfo{journal}{\emph{Advances in neural information processing
  systems}}  \bibinfo{volume}{25} (\bibinfo{year}{2012}),
  \bibinfo{pages}{1097--1105}.
\newblock


\bibitem[\protect\citeauthoryear{LeCun, Haffner, Bottou, and Bengio}{LeCun
  et~al\mbox{.}}{1999}]%
        {lecun1999object}
\bibfield{author}{\bibinfo{person}{Yann LeCun}, \bibinfo{person}{Patrick
  Haffner}, \bibinfo{person}{L{\'e}on Bottou}, {and} \bibinfo{person}{Yoshua
  Bengio}.} \bibinfo{year}{1999}\natexlab{}.
\newblock \showarticletitle{Object recognition with gradient-based learning}.
\newblock In \bibinfo{booktitle}{\emph{Shape, contour and grouping in computer
  vision}}. \bibinfo{publisher}{Springer}, \bibinfo{pages}{319--345}.
\newblock


\bibitem[\protect\citeauthoryear{Leroy and Goedem{\'e}}{Leroy and
  Goedem{\'e}}{2021}]%
        {leroy2021optimal}
\bibfield{author}{\bibinfo{person}{Paul~Albert Leroy} {and}
  \bibinfo{person}{Toon Goedem{\'e}}.} \bibinfo{year}{2021}\natexlab{}.
\newblock \showarticletitle{Optimal Distribution of CNN Computations on Edge
  and Cloud.}. In \bibinfo{booktitle}{\emph{ICAART (2)}}.
  \bibinfo{pages}{604--613}.
\newblock


\bibitem[\protect\citeauthoryear{Li, Zhou, and Chen}{Li et~al\mbox{.}}{2018}]%
        {li2018edge}
\bibfield{author}{\bibinfo{person}{En Li}, \bibinfo{person}{Zhi Zhou}, {and}
  \bibinfo{person}{Xu Chen}.} \bibinfo{year}{2018}\natexlab{}.
\newblock \showarticletitle{Edge intelligence: On-demand deep learning model
  co-inference with device-edge synergy}. In
  \bibinfo{booktitle}{\emph{Proceedings of the 2018 Workshop on Mobile Edge
  Communications}}. \bibinfo{pages}{31--36}.
\newblock


\bibitem[\protect\citeauthoryear{Louis, Azad, Delshadtehrani, Gupta, Warden,
  Reddi, and Joshi}{Louis et~al\mbox{.}}{2019}]%
        {louis2019towards}
\bibfield{author}{\bibinfo{person}{Marcia~Sahaya Louis}, \bibinfo{person}{Zahra
  Azad}, \bibinfo{person}{Leila Delshadtehrani}, \bibinfo{person}{Suyog Gupta},
  \bibinfo{person}{Pete Warden}, \bibinfo{person}{Vijay~Janapa Reddi}, {and}
  \bibinfo{person}{Ajay Joshi}.} \bibinfo{year}{2019}\natexlab{}.
\newblock \showarticletitle{Towards deep learning using tensorFlow lite on
  RISC-V}. In \bibinfo{booktitle}{\emph{Third Workshop on Computer Architecture
  Research with RISC-V (CARRV)}}, Vol.~\bibinfo{volume}{1}. \bibinfo{pages}{6}.
\newblock


\bibitem[\protect\citeauthoryear{Matsubara, Baidya, Callegaro, Levorato, and
  Singh}{Matsubara et~al\mbox{.}}{2019}]%
        {matsubara2019distilled}
\bibfield{author}{\bibinfo{person}{Yoshitomo Matsubara}, \bibinfo{person}{Sabur
  Baidya}, \bibinfo{person}{Davide Callegaro}, \bibinfo{person}{Marco
  Levorato}, {and} \bibinfo{person}{Sameer Singh}.}
  \bibinfo{year}{2019}\natexlab{}.
\newblock \showarticletitle{Distilled split deep neural networks for
  edge-assisted real-time systems}. In \bibinfo{booktitle}{\emph{Proceedings of
  the 2019 Workshop on Hot Topics in Video Analytics and Intelligent Edges}}.
  \bibinfo{pages}{21--26}.
\newblock


\bibitem[\protect\citeauthoryear{Mehta and Shorey}{Mehta and Shorey}{2020}]%
        {mehta2020deepsplit}
\bibfield{author}{\bibinfo{person}{Rishabh Mehta} {and} \bibinfo{person}{Rajeev
  Shorey}.} \bibinfo{year}{2020}\natexlab{}.
\newblock \showarticletitle{DeepSplit: Dynamic Splitting of Collaborative
  Edge-Cloud Convolutional Neural Networks}. In \bibinfo{booktitle}{\emph{2020
  International Conference on COMmunication Systems \& NETworkS (COMSNETS)}}.
  IEEE, \bibinfo{pages}{720--725}.
\newblock


\bibitem[\protect\citeauthoryear{pytorch}{pytorch}{[n.d.]}]%
        {torchhub}
\bibfield{author}{\bibinfo{person}{pytorch}.}
  \bibinfo{year}{[n.d.]}\natexlab{}.
\newblock \bibinfo{booktitle}{\emph{Pytorch Hub}}.
\newblock
\urldef\tempurl%
\url{https://pytorch.org/hub/}
\showURL{%
Retrieved July 2021 from \tempurl}


\bibitem[\protect\citeauthoryear{raspberrypi}{raspberrypi}{[n.d.]}]%
        {rpi}
\bibfield{author}{\bibinfo{person}{raspberrypi}.}
  \bibinfo{year}{[n.d.]}\natexlab{}.
\newblock \bibinfo{booktitle}{\emph{Raspberry Pi 4 Model}}.
\newblock
\urldef\tempurl%
\url{https://www.raspberrypi.org/products/raspberry-pi-4-model-b/}
\showURL{%
Retrieved July 2021 from \tempurl}


\bibitem[\protect\citeauthoryear{Sandler, Howard, Zhu, Zhmoginov, and
  Chen}{Sandler et~al\mbox{.}}{2018}]%
        {sandler2018mobilenetv2}
\bibfield{author}{\bibinfo{person}{Mark Sandler}, \bibinfo{person}{Andrew
  Howard}, \bibinfo{person}{Menglong Zhu}, \bibinfo{person}{Andrey Zhmoginov},
  {and} \bibinfo{person}{Liang-Chieh Chen}.} \bibinfo{year}{2018}\natexlab{}.
\newblock \showarticletitle{Mobilenetv2: Inverted residuals and linear
  bottlenecks}. In \bibinfo{booktitle}{\emph{Proceedings of the IEEE conference
  on computer vision and pattern recognition}}. \bibinfo{pages}{4510--4520}.
\newblock


\bibitem[\protect\citeauthoryear{Simonyan and Zisserman}{Simonyan and
  Zisserman}{2014}]%
        {simonyan2014very}
\bibfield{author}{\bibinfo{person}{Karen Simonyan} {and}
  \bibinfo{person}{Andrew Zisserman}.} \bibinfo{year}{2014}\natexlab{}.
\newblock \showarticletitle{Very deep convolutional networks for large-scale
  image recognition}.
\newblock \bibinfo{journal}{\emph{arXiv preprint arXiv:1409.1556}}
  (\bibinfo{year}{2014}).
\newblock


\bibitem[\protect\citeauthoryear{Szegedy, Ioffe, Vanhoucke, and Alemi}{Szegedy
  et~al\mbox{.}}{2017}]%
        {szegedy2017inception}
\bibfield{author}{\bibinfo{person}{Christian Szegedy}, \bibinfo{person}{Sergey
  Ioffe}, \bibinfo{person}{Vincent Vanhoucke}, {and}
  \bibinfo{person}{Alexander~A Alemi}.} \bibinfo{year}{2017}\natexlab{}.
\newblock \showarticletitle{Inception-v4, inception-resnet and the impact of
  residual connections on learning}. In \bibinfo{booktitle}{\emph{Thirty-first
  AAAI conference on artificial intelligence}}.
\newblock


\bibitem[\protect\citeauthoryear{Tang, Chen, Zeng, Yu, and Chen}{Tang
  et~al\mbox{.}}{2020}]%
        {tang2020joint}
\bibfield{author}{\bibinfo{person}{Xin Tang}, \bibinfo{person}{Xu Chen},
  \bibinfo{person}{Liekang Zeng}, \bibinfo{person}{Shuai Yu}, {and}
  \bibinfo{person}{Lin Chen}.} \bibinfo{year}{2020}\natexlab{}.
\newblock \showarticletitle{Joint multi-user DNN partitioning and computational
  resource allocation for collaborative edge intelligence}.
\newblock \bibinfo{journal}{\emph{IEEE Internet of Things Journal}}
  (\bibinfo{year}{2020}).
\newblock


\bibitem[\protect\citeauthoryear{ubuntu}{ubuntu}{[n.d.]}]%
        {ubuntu}
\bibfield{author}{\bibinfo{person}{ubuntu}.} \bibinfo{year}{[n.d.]}\natexlab{}.
\newblock \bibinfo{booktitle}{\emph{Ubuntu Desktop}}.
\newblock
\urldef\tempurl%
\url{https://ubuntu.com/download/desktop}
\showURL{%
Retrieved July 2021 from \tempurl}


\bibitem[\protect\citeauthoryear{van~der Meulen}{van~der Meulen}{[n.d.]}]%
        {gartner}
\bibfield{author}{\bibinfo{person}{Rob van~der Meulen}.}
  \bibinfo{year}{[n.d.]}\natexlab{}.
\newblock \bibinfo{booktitle}{\emph{What Edge Computing Means for
  Infrastructure and Operations Leaders}}.
\newblock
\urldef\tempurl%
\url{https://www.gartner.com/smarterwithgartner/what-edge-computing-means-for-infrastructure-and-operations-leaders/}
\showURL{%
Retrieved July 2021 from \tempurl}


\bibitem[\protect\citeauthoryear{Zhou, Wen, Teodorescu, and Du}{Zhou
  et~al\mbox{.}}{2019}]%
        {zhou2019distributing}
\bibfield{author}{\bibinfo{person}{Li Zhou}, \bibinfo{person}{Hao Wen},
  \bibinfo{person}{Radu Teodorescu}, {and} \bibinfo{person}{David~HC Du}.}
  \bibinfo{year}{2019}\natexlab{}.
\newblock \showarticletitle{Distributing deep neural networks with
  containerized partitions at the edge}. In \bibinfo{booktitle}{\emph{2nd
  $\{$USENIX$\}$ Workshop on Hot Topics in Edge Computing (HotEdge 19)}}.
\newblock


\end{thebibliography}

\end{document}